# Online Robust Sliding-Windowed LiDAR SLAM in Natural Environments

Quang-Ha Pham, Ngoc-Huy Tran[✉], Thanh-Toan Nguyen, Thien-Phuc Tran

*Abstract*— Despite the growing interest for autonomous environmental monitoring, effective SLAM realization in native habitats remains largely unsolved. In this paper, we fill this gap by presenting a novel online graph-based SLAM system for 2D LiDAR sensor in natural environments. By taking advantage of robust weighting scheme, sliding-windowed optimization, fast scan-matcher and parallel computing, our system not only delivers stable performance in cluttered surroundings but also meets real-time constraint. Simulated and experimental results confirm the feasibility and efficiency in the overall design of the proposed system.

*Keywords—SLAM, robust kernel, sliding window, scan matching, parallel computing, natural environment*

## I. Introduction

SLAM is considered as one of key technologies in fully-autonomous vehicles. A robot capable of mapping and self-localization concurrently can perform online path planning, avoid obstacles and reduce accumulated localized error caused by dead-reckoning system [1]. Not only has SLAM been deployed in different environments (indoor [2], urban [3], riverine [4], seabed [5]) but it has been realized by various modalities (LiDAR [2], camera [6], sonar [5], radar [4]).

Application of 2D LiDAR SLAM systems in man-made environments is a story of great success [7] [8] [2]. However, their viability in natural ones (forests, rivers) remains questionable. It is apparent that traditional SLAM systems degrade deeply in quality due to cluttered scene, deficiency of distinguished features (corners, edges) and abrupt illumination variation [9]. Another reason that hinders the development of an efficient SLAM system in those environments is a shortage of useful site-specific datasets: [9] only provides raw camera images, [10] does records 2D LiDAR scans but lacks ground truth. Fortunately, recent success of VO system (sub-problem of visual SLAM) in riverine habitat [11] sheds light on how to tackle some challenges faced by current SLAM systems. To account for global and local lighting changes, the author proves that an adaptive weighting strategy makes cost function more illumination-invariant. Thus, robust weighting kernels show promising potential in contributing to a workable SLAM solution for arbitrary modality in general.

Nowadays, backed by rocketing enhancement in microprocessor power, memory capacity, computation algorithm, graph optimization based on nonlinear least-squares become the mainstream of SLAM research. Traditionally, most 2D LiDAR SLAM systems rely on time-consuming batch processing, which optimizes a whole graph structure at a time, for offline map reconstruction [12] [7]. For online mapping, sliding-windowed approach is usually exploited to maintain constant update time in visual SLAM [6] [13]. However, this method is seldomly implemented in the 2D LiDAR SLAM literature, which partly stems from slow correlative scan matching techniques [14] [15]. A common alternative is submapping, which is employed by Cartographer [2], but this introduces cumbersome map management system and impedes sensor fusion in the long run. Therefore, adopting a fast scan-matcher is essential to bring sliding-windowed 2D LiDAR SLAM into practice.

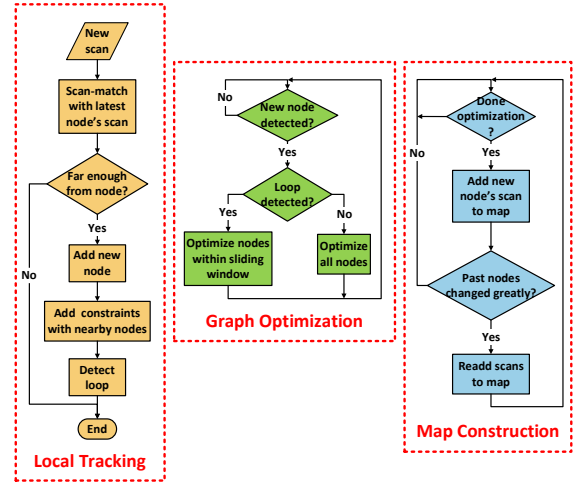

Fig. 1. Block diagram of the overall SLAM system: while *Local Tracking* interrupts to process raw LiDAR scans, other two modules constantly check for relevant flags to carry out their tasks.

Our main contribution in this paper is to design a novel 2D LiDAR SLAM that leverages robust weighting scheme, sliding-windowed optimization, fast scan-matcher and parallel computing to exhibit stable and real-time performance in natural environments. Similar to [2], we utilize the power of parallelization by dividing the system into three building modules, each of which runs in separate thread (Fig. 1).

- *Local Tracking* estimates LiDAR pose and updates graph structure. Inspired by [6], a newly created node is connected to nearby ones to ensure the system's robustness. However, multiple scan alignments in tens of milliseconds is impossible unless a fast scan-matcher is used. Thus, we adopt Andrea's PLICP [16] thanks to its superior speed over correlative methods. To finish the proof of Andrea, we expand his quartic equation to obtain an explicit formula for every coefficient. In addition, to estimate uncertainty of the matching result, we apply a theorem from [17] to 2D LiDAR measurement and arrive at a closed-formed solution for uncertainty covariance.

- *Graph Optimization* optimizes partial or entire graph structure. When a node is created, all nodes in the current sliding window are optimized, thereby reducing the ill-effect of local optima in past optimization steps. To prevent outliers from deteriorating the system, every consecutive graph constraint is robustified by a generalized robust kernel. Loop is detected based on distance metrics to eliminate drift and maintain global map consistency.

- *Map Construction* manages the insertion and removal of scans in the map. We choose occupancy grid map thanks to its straightforward

implementation but wide application in other robotic tasks such as path planning, obstacle avoidance.

Simulated results confirm that our SLAM system can operate in real-time (execution time of each thread fluctuates around 15ms), make small error over long travelling distance (ATE of 0.6m in a 500-metered run). Experimental results further demonstrate the system's robustness under cluttered scene around a lake. This paper is structured as followed: section II to IV give detailed description of each module's operational principle, section V analyses the system's performance through simulated and experimental results, section VI concludes and discusses further applications.

## II. LOCAL TRACKING

### A. Scan Matching

The aim of scan matching is to find a rigid transformation that aligns two scans: given a reference surface $\mathcal{S}^{\text{ref}}$ approximated by reference scanpoints $\{\mathbf{p}_j\}$, find a roto-translation $\mathbf{q} = [\mathbf{t} \ \theta]^T$ that minimizes the distance from each $\mathbf{p}_i^w = \mathbf{p}_i \oplus \mathbf{q}$ in current scanpoints $\{\mathbf{p}_i\}$ to its projection $\prod\{\mathcal{S}^{\text{ref}}, \mathbf{p}_i^w\}$ on surface $\mathcal{S}^{\text{ref}}$. Among various algorithms, Andrea's PLICP [16] is chosen thanks to its rapid and finite convergence properties. According to Andrea, for every $\mathbf{p}_i^w$, a corresponding search is carried out to find two closest reference points in $\{\mathbf{p}_j\}$, denoted $\mathbf{p}_{j_1^i}$ and $\mathbf{p}_{j_2^i}$. Without loss of generality, set $\mathbf{p}_{j_1^i} = \prod\{\mathcal{S}^{\text{ref}}, \mathbf{p}_i^w\}$ and the normal vector $\mathbf{n}_i$ of $\mathcal{S}^{\text{ref}}$ at $\mathbf{p}_{j_1^i}$ is perpendicular to $\mathbf{p}_{j_1^i} - \mathbf{p}_{j_2^i}$. PLICP is formulated as:

$$\mathbf{q}_{k+1} = \arg\min \sum_i \left(\mathbf{n}_i^T \left[\mathbf{p}_i \oplus \mathbf{q}_{k+1} - \mathbf{p}_{j_1^i}\right]\right)^2 \quad (1)$$

Andrea proves that (1) possesses a closed-formed solution that relates to solution of a quartic equation with unknown $\lambda$:

$$\mathbf{g}^T (2\mathbf{M} + 2\lambda\mathbf{W})^{-1} \mathbf{W} (2\mathbf{M} + 2\lambda\mathbf{W})^{-T} \mathbf{g} = 1 \quad (2)$$

For completion, we expand (2) to obtain those coefficients:

$$a\lambda^4 + b\lambda^3 + c\lambda^2 + d\lambda + e = 0 \quad (3)$$

- $a = 16$
- $b = 16 \text{tr}(\mathbf{S})$
- $c = 8|\mathbf{S}| + 4\text{tr}(\mathbf{S})^2 - 4\mathbf{g}^T \begin{bmatrix} \mathbf{A}^{-1}\mathbf{B}\mathbf{B}^T\mathbf{A}^{-T} & -\mathbf{A}^{-1}\mathbf{B} \\ -\mathbf{A}^{-1}\mathbf{B} & 1 \end{bmatrix} \mathbf{g}$
- $d = 4|\mathbf{S}|\text{tr}(\mathbf{S}) - 4\mathbf{g}^T \begin{bmatrix} \mathbf{A}^{-1}\mathbf{B}\mathbf{S}^A\mathbf{B}^T\mathbf{A}^{-T} & -\mathbf{A}^{-1}\mathbf{B}\mathbf{S}^A \\ -\mathbf{A}^{-1}\mathbf{B}\mathbf{S}^A & \mathbf{S}^A \end{bmatrix} \mathbf{g}$
- $e = |\mathbf{S}|^2 - \mathbf{g}^T \begin{bmatrix} \mathbf{A}^{-1}\mathbf{B}\mathbf{S}^{A^T}\mathbf{S}^A\mathbf{B}^T\mathbf{A}^{-T} & -\mathbf{A}^{-1}\mathbf{B}\mathbf{S}^{A^T}\mathbf{S}^A \\ -\mathbf{A}^{-1}\mathbf{B}\mathbf{S}^{A^T}\mathbf{S}^A & \mathbf{S}^{A^T}\mathbf{S}^A \end{bmatrix} \mathbf{g}$

### B. Uncertainty Estimation

To predict uncertainty of an arbitrary estimation, in computer vision communities, the following theorem [17] is often used: let $\mathbf{x}$ be the result of minimizing a cost function $J$ which depends on some $\mathbf{z}$ measurement, then it follows that the uncertainty covariance can be approximated by:

$$\text{cov}(\mathbf{x}) \cong \frac{\partial^2 J}{\partial \mathbf{x}^2}^{-1} \frac{\partial^2 J}{\partial \mathbf{x} \partial \mathbf{z}} \text{cov}(\mathbf{z}) \frac{\partial^2 J}{\partial \mathbf{x} \partial \mathbf{z}}^T \frac{\partial^2 J}{\partial \mathbf{x}^2}^{-1} \quad (4)$$

In our scan matching's case, $\mathbf{x} = [\mathbf{t} \ \theta]^T$ and the cost function is:

$$J = \sum_i \left(\mathbf{n}_i^T \left[(\mathbf{R}(\theta)\mathbf{p}_i + \mathbf{t}) - \mathbf{p}_{j_1^i}\right]\right)^2 = \sum_i J_i \quad (5)$$

For 2D LiDAR, $\text{cov}(\mathbf{z})$ corresponds to the standard deviation of range measurement, which is mentioned in datasheet. To facilitate the following deduction, every scanpoint $\mathbf{p}_i$ and $\mathbf{p}_j$ is converted to its equivalent radial representation $\mathbf{p}_i = \rho_i \vartheta_i$ and $\mathbf{p}_j = \rho_j \vartheta_j$ with known directional vector $\vartheta$ and measured range $\rho$. In formulae, $\mathbf{z}$ is formed by concatenating the current scanpoint with its two closest reference ones:

$$\mathbf{z} = [\rho_1 \ \rho_{j_1^1} \ \rho_{j_2^1} \ \cdots \ \rho_i \ \rho_{j_1^i} \ \rho_{j_2^i} \ \cdots]^T \quad (6)$$

Following Andrea, we set $\mathbf{C}_i = \mathbf{n}_i \mathbf{n}_i^T$, $\mathbf{v}_1 = \mathbf{R}\left(\theta + \frac{\pi}{2}\right)\mathbf{p}_i$, $\mathbf{v}_2 = \mathbf{R}(\theta)\mathbf{p}_i + \mathbf{t} - \mathbf{p}_{j_1^i}$, $\mathbf{v}_3 = \mathbf{R}(\theta)\vartheta_i$, $\mathbf{v}_4 = \mathbf{R}\left(\theta + \frac{\pi}{2}\right)\vartheta_i$, $\mathbf{v}_5 = \mathbf{R}(\theta + \pi)\mathbf{p}_i$. Taking second partial derivatives, we clarify his proof:

$$\frac{\partial^2 J_i}{\partial \mathbf{x}^2} = \begin{bmatrix} 2\mathbf{C}_i & 2\mathbf{C}_i \mathbf{v}_1 \\ 2\mathbf{v}_1^T \mathbf{C}_i & 2\mathbf{v}_2^T \mathbf{C}_i \mathbf{v}_5 + 2\mathbf{v}_1^T \mathbf{C}_i \mathbf{v}_1 \end{bmatrix} \quad (7)$$

$$\frac{\partial^2 J_i}{\partial \mathbf{x} \partial \mathbf{z}} = \begin{bmatrix} 0 & 0 & 0 & \cdots \\ 0 & 0 & 0 & \cdots \end{bmatrix} \quad (8)$$

$$\begin{bmatrix} 2\mathbf{C}_i \mathbf{v}_3 & -2\mathbf{C}_i \vartheta_{j_1^i} + 2\frac{\partial \mathbf{C}_i}{\partial \rho_{j_1^i}} \mathbf{v}_2 & 2\frac{\partial \mathbf{C}_i}{\partial \rho_{j_2^i}} \mathbf{v}_2 & \cdots \\ 2\mathbf{v}_2^T \mathbf{C}_i \mathbf{v}_4 + 2\mathbf{v}_3^T \mathbf{C}_i \mathbf{v}_1 & 2\mathbf{v}_2^T \frac{\partial \mathbf{C}_i}{\partial \rho_{j_1^i}} \mathbf{v}_1 - 2\mathbf{v}_1^T \mathbf{C}_i \vartheta_{j_1^i} & 2\mathbf{v}_2^T \frac{\partial \mathbf{C}_i}{\partial \rho_{j_2^i}} \mathbf{v}_1 & \cdots \end{bmatrix}$$

with $\epsilon = 0.001$:

$$\frac{\partial \mathbf{C}_i}{\partial \rho_{j_{1,2}^i}} \cong \frac{\mathbf{C}_i\left(\rho_{j_{1,2}^i} + \epsilon.\text{sgn}(\rho_{j_{1,2}^i})\|\rho_{j_{1,2}^i}\|\right) - \mathbf{C}_i\left(\rho_{j_{1,2}^i}\right)}{\epsilon} \quad (9)$$

Summing all individual derivatives, we get:

$$\frac{\partial^2 J}{\partial \mathbf{x}^2} = \sum_i \frac{\partial^2 J_i}{\partial \mathbf{x}^2}, \quad \frac{\partial^2 J}{\partial \mathbf{x} \partial \mathbf{z}} = \sum_i \frac{\partial^2 J_i}{\partial \mathbf{x} \partial \mathbf{z}} \quad (10)$$

### C. Module Details

When a LiDAR scan is received, the scan-matcher aligns it with the latest node's scan to get LiDAR relative displacement, which is subsequently concatenated with the latest node's pose to get LiDAR current pose. We set the previous roto-translation as initial guess to boost convergence rate of the scan-matcher. If the matching result exceeds a predefined threshold in translation or rotation, a new node is created with its pose being the current LiDAR pose. Since scan matching is an iterative procedure, global convergence is not guaranteed. Therefore, to prevent the whole system from collapse, the newly created node is densely connected with nearby ones in a predefined radius, hoping subsequent optimization steps pull past nodes closer to their global optima (Fig. 2). Loop closing mechanism, in our system, is simple: we create a loop constraint between two nodes whenever their indexes differ from each other by a certain threshold.

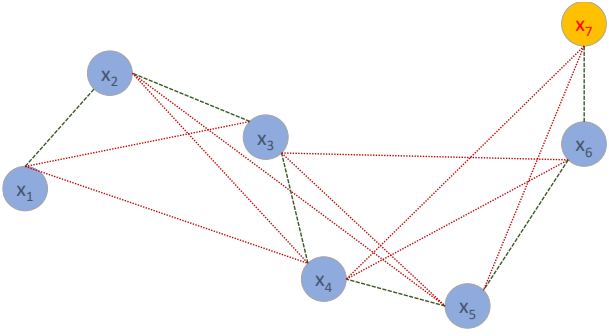

Fig. 2. Illustration of dense pose graph construction: each red line represents an extra connection between the current node and a nearby node within predefined radius.

### III. GRAPH OPTIMIZATION

#### A. Robust Kernel

*Nonlinear least-squares* (NLS) is frequently used for real-time graph optimization. However, NLS can get stuck in local optima if data contain many outliers. This is especially the case of online SLAM that operates in sophisticated environments. To relieve the ill-effect of outliers, one of the most popular methods is robust kernels, which flatten large residuals to minimize their influence on the solution. The NLS combined with robust kernel can be solved by the *Iterative Reweighted Least Squares* (IRLS) [18] approach.

Let $\mathbf{x} = [\mathbf{x}_1^T \ ... \ \mathbf{x}_n^T]^T$ be vector of all LiDAR poses from $t_1$ to $t_n$ with $\mathbf{x}_i = [x \ y \ \theta]^T$. The purpose of IRLS is to estimate $\mathbf{x}$:

$$\mathbf{x}^* = \arg\min \sum_{i,j} w_{ij} r_{ij}(\mathbf{x})^2 \quad (11)$$

by minimizing the total sum of weighted squared residuals $r_{ij}(\mathbf{x})$:

$$r_{ij}(\mathbf{x}) = \sqrt{\mathbf{e}_{ij}(\mathbf{x})^T \mathbf{\Omega}_{ij} \mathbf{e}_{ij}(\mathbf{x})} \quad (12)$$

with $\mathbf{e}_{ij}(\mathbf{x}) = \hat{\mathbf{z}}_{ij}(\mathbf{x}_i, \mathbf{x}_j) - \mathbf{z}_{ij}$ be the error between $\hat{\mathbf{z}}_{ij} = \mathbf{x}_i - \mathbf{x}_j$ and the estimated roto-translation $\mathbf{z}_{ij}$, uncertainty covariance $\mathbf{\Omega}_{ij}$ from the scan-matcher. By utilizing robust kernel, each residual's weight is adaptively computed based on the residual's magnitude:

$$w_{ij} = \frac{1}{r_{ij}(\mathbf{x})} \cdot \rho'\left(r_{ij}(\mathbf{x})\right) \quad (13)$$

with $\rho(r)$ be the characteristic function of the robust kernel. For convenience, we shall implement Barron's kernel that generalizes several kernels [19]:

$$\rho(r, \alpha, c) = \frac{|\alpha - 2|}{\alpha}\left(\left(\frac{\left(\frac{r}{c}\right)^2}{|\alpha - 2|} + 1\right)^{\frac{\alpha}{2}} - 1\right) \quad (14)$$

By simple tuning $\alpha$, we can interpolate between various robust kernels: L2 ($\alpha = 2$), L1 ($\alpha = 1$), Cauchy ($\alpha = 0$), Geman-McClure ($\alpha = -2$), Welsh ($\alpha = -\infty$). In this paper, $\alpha$ is manually selected to deal with different kinds of outliers. A system of dynamically adapting $\alpha$ is a promising research area yet to be explored.

#### B. Module Details

When a new node is detected, it is immediately added to the current sliding window. All consecutive constraints are robustified because they possess higher potential of being outliers. Optimization then takes place within those nodes residing in the current sliding window, thereby ensuring constant-time graph update. Upon detecting a loop, to eliminate drift and maintain global map consistency, the whole graph is optimized. At first glance, this may seen too time-consuming, but it worths noting that graph optimization runs in a separate thread, and loop closure is not a frequent occurrence, so batch optimization in this case is acceptable.

### IV. MAP CONSTRUCTION

#### A. Occupancy Grid Map Update

A common map type for 2D SLAM is occupancy grid map: each cell $i$ is either marked occupied ($m_i$) or empty ($\neg m_i$). To represent the probabilistic relationship between measurement's history $\mathbf{z}_{1:t}$, agent's trajectory $\mathbf{x}_{1:t}$ and cell state $\{m_i\}$, *Hidden Markov Model* (HMM) [20] is usually applied: $\mathbf{z}_t$ only depends on $\mathbf{x}_t$, $\{m_i\}$ is independent of $\mathbf{x}_t$, each $m_i$ is pairwise independent of one another. Also in [20], the log-odds update rule for occupancy grid map, in formulae, is:

$$l(m_i|\mathbf{z}_{1:t}, \mathbf{x}_{1:t}) = l(m_i|\mathbf{z}_{1:t-1}, \mathbf{x}_{1:t-1}) \quad (15)$$
$$+ l(m_i|\mathbf{z}_t, \mathbf{x}_t) - l(m_i)$$

In case of 2D LiDAR, the measurement is in the form of $\mathbf{z}_t = [\rho_1 \ \rho_2 \ \cdots \ \rho_N]$ with $\rho_i$ be the reflected distance from a surface at angle $i$ and $N$ be number of rays per sampling interval of LiDAR. To calculate $p(m_i|\mathbf{z}_t, \mathbf{x}_t)$, firstly, each ray in $\mathbf{z}_t$ is discretized to the map's resolution by Bresenham's algorithm [21]. Then, cell at the final end is considered occupied and the other is considered empty:

$$p(m_i|\mathbf{z}_t, \mathbf{x}_t) = \begin{cases} prop\_occupied, & m_i \\ prop\_free, & \neg m_i \end{cases} \quad (16)$$

#### B. Module Details

Upon witnessing a successful optimization step, the latest scan is added to the current map by means of updating the occupancy probability of each cell along individual ray. Also, past scans can be read if their related nodes deviate much in pose since the last update. This simple procedure, however, does not guarantee optimal map retrieval at any time, but is applicable to other robotic tasks as long as the map resolution is kept above a minimum acceptable value.

### V. RESULTS

#### A. Implementation

*Robot Operating System* (ROS) is a popular middleware in the implementation of many robotic tasks. In this paper, ROS is used to collect LiDAR scans, exchange messages among different modules and visualize algorithmic results. To solve Andrea's quartic equation in the scan-matcher, we employ Eigen library. We use g2o [22] to manage graph structure and choose Levenberg-Marquardt algorithm with a maximum of 100 steps to solve IRLS. ATE and RPE metrics [23] are selected for quantitative evaluation of localized error.

#### B. Simulation

In Gazebo environment, we drive a USV around a valley of size 200x200m (Fig. 3). USV is equipped with a 2D LiDAR

sensor (coverage angle of 270º, angular resolution of 0.25º, maximum range of 60m, scanning rate of 40Hz). In correlation with the environment size, we set the map's resolution to 0.4m (Fig. 5) for online mapping. A map of higher resolution (0.1m) (Fig. 4) can be generated offline after travelling. All simulated results are implemented in Intel Core i7-3770 3.40GHz. Statistics are summarized in Table I.

At the end of the journey, when USV returns to its departure point, loop is detected, multiple loop constraints are created, all nodes in the graph is optimized and many scans are read, making execution time each thread escalates. As an effect of loop closure, accumulated localized error is significantly reduced, making the map more globally consistent (Fig. 8).

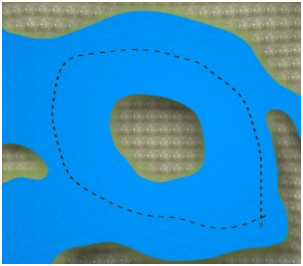 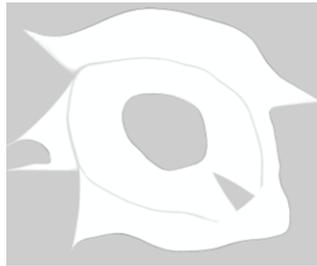

Fig. 3. Simulated valley and the vehicle's trajectory       Fig. 4. Complete map of valley

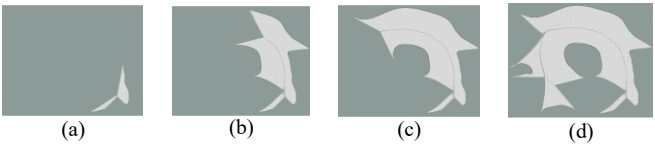

Fig. 5. Map reconstruction step at 7 (a), 167 (b), 297 (c) and 450 (d) nodes.

Looking at Fig. 6, it is apparent that the execution time of each thread fluctuates around a constant level, thus confirming real-time capability of the whole system.

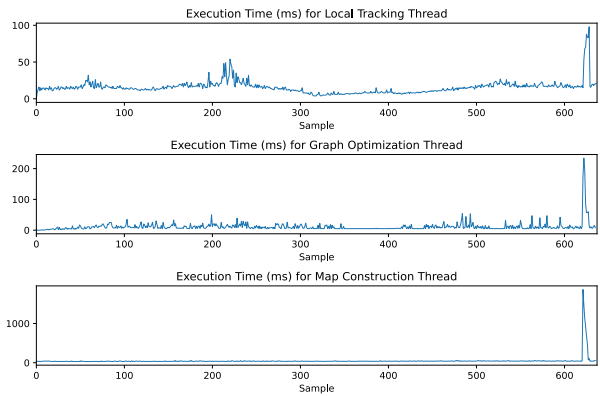

Fig. 6. Execution time for three threads.

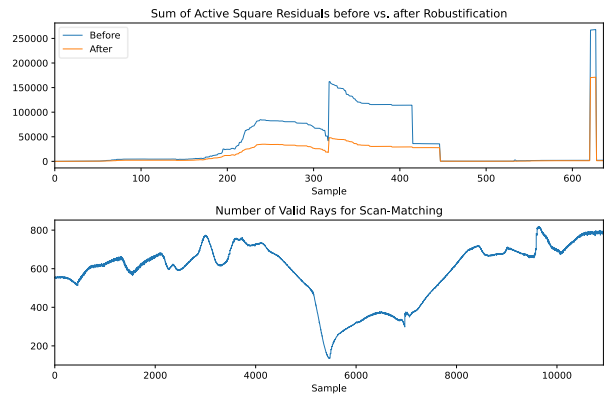

Fig. 7. Sum of active square residuals (above) vs. number of valid rays for scan matching (below).

When entering a wide area, the number of valid rays for scan matching declines sharply, making the estimation's uncertainty increase. As a result, the scan-matcher is more likely to get stuck in local optima. This observation is verified by some sudden rises in sum of active squared residuals (Fig. 7). We can see the impact of robust kernel in this case in reducing the magnitude of large residuals that threaten to shut down the system.

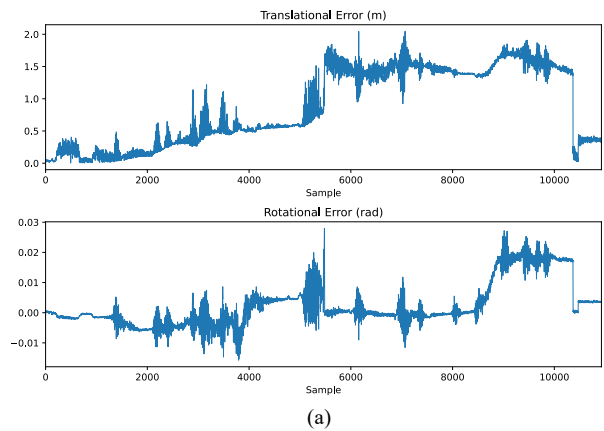

(a)

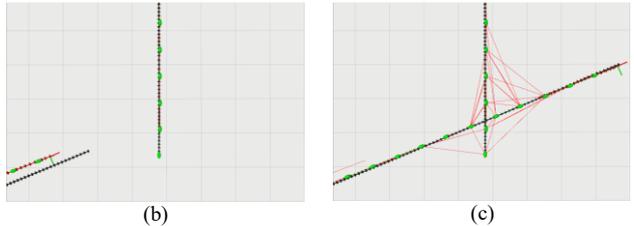

(b)    (c)

Fig. 8. Localization before (b) and after loop closure (c). After closing loop, the estimated pose (red) approaches ground truth (black), thus eliminating drift after the long run (a).

Since our SLAM system consists of many concurrent tasks, each of which gains random access to the computer resources, randomness in results is unavoidable. Thus, to make the results more reliable, especially for ATE and RPE, we repeat the simulation 10 times (Fig. 9), then plot the values of ATE and RPE for each attempt to examine their variation. As can be seen from Fig. 9, ATE and RPE fluctuate around 0.61m and 0.21m respectively with the largest deviation of 0.02m, thereby confirming the system's reliability despite the fact that there exists some randomness in the system's operation.

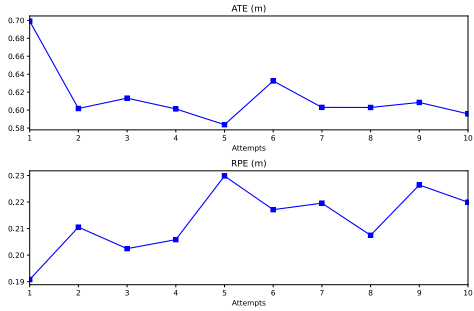

Fig. 9. Evaluation of the system's reliability after 10 attempts.

TABLE I.    SIMULATION STATISTICS

| | |
|---|---|
| Number of graph nodes | 638 |
| Number of graph constraints | 1952 |
| Travelling distance | 496.728 m |
| Execution time for scan matching (avg – max) | 10 – 35 ms |
| Execution time for local tracking (avg – max) | 15 – 98 ms |
| Execution time for graph optimization (avg – max) | 11 – 234 ms |
| Execution time for map construction (avg – max) | 49 – 1866 ms |
| RPE | 0.217 m |
| ATE | 0.632 m |

*C. Experiment*

USV2000 (Fig. 11) is a new generation of autonomous surface vehicles that leverages two rear thrusters for moving forward and two side ones for turning. Jetson Nano embedded computer plays a central role in realizing control and navigation algorithms. Each program in the computer is packaged in a ROS node to take advantage of efficient data interchange in the ROS ecosystem. The embedded computer communicates with STM32F407 microcontrolller by CAN bus and with VIAM-USV-QG ground control station by Wifi network. To realize our SLAM system, we equip USV2000 with Hokuyo UTM-30LX 2D LiDAR sensor (coverage angle of 270°, angular resolution of 0.25°, maximum range of 39m, scanning rate of 40Hz). In this experiment, the surface vehicle is commanded to travel a short journey close to the bank of Phu Tho Lake (Fig. 10). Statistics are summarized in Table II.

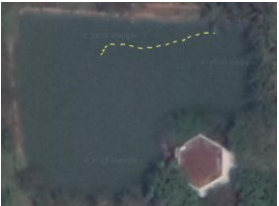
Fig. 10. Experimental location and the vehicle's trajectory

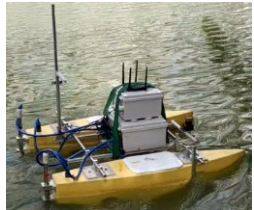
Fig. 11. USV2000

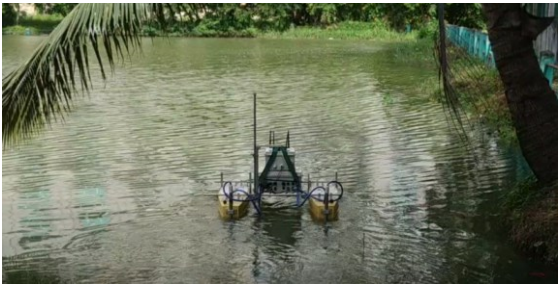
Fig. 12. Phu Tho Lake.

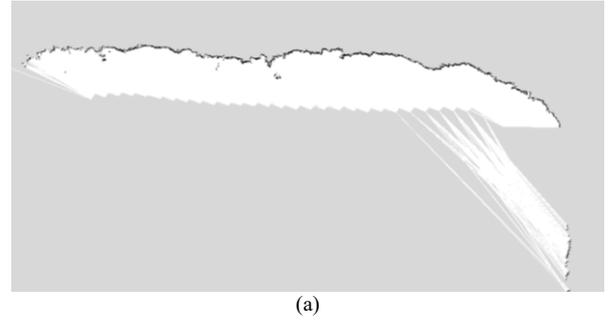
(a)

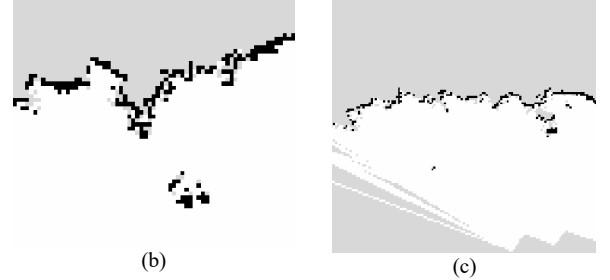
(b)                                    (c)

Fig. 13. Partial map of the lake (a) and zoom-in some parts (b-c).

Looking at Fig. 14, it is apparent that the execution time of each thread fluctuates around a constant level, thus confirming real-time capability of the whole system.

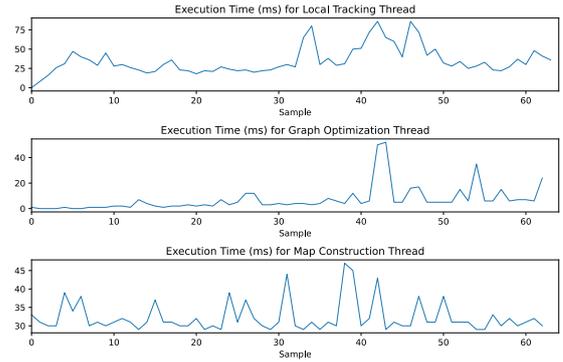

Fig. 14. Execution time for three threads.

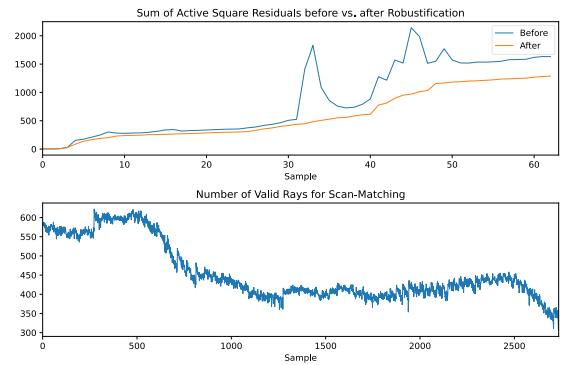

Fig. 15. Sum of active square residuals (above) vs. number of valid rays for scan matching (below).

Looking at the generated map (Fig. 13) with a resolution of 5cm, it worths emphasizing that although the surroundings are cluttered with leaves and bushes, our SLAM system is able to robustly create a detailed description of the riverine scenery. To further examine the system's robustness, looking at Fig. 15, it is noticeable that although the number of valid rays for scan matching decreases substantially as USV moves farther from the bank, robust kernel is able to withstand the effect of rapid increase in sum of active squared residuals, thus maintaining the system's stability throughout the journey.

TABLE II. EXPERIMENT STATISTICS

| | |
|---|---|
| Number of graph nodes | 65 |
| Number of graph constraints | 330 |
| Travelling distance | 32.082 m |
| Execution time for scan matching (avg – max) | 10 – 22 ms |
| Execution time for local tracking (avg – max) | 35 – 86 ms |
| Execution time for graph optimization (avg – max) | 7 – 52 ms |
| Execution time for map construction (avg – max) | 32 – 47 ms |
| RPE | - |
| ATE | - |

## VI. CONCLUSIONS

In this paper, our proposed graph-based SLAM that utilizes 2D LiDAR as the unique modality is designed with the philosophy of modularization and parallelization in order to simultaneously estimating the LiDAR pose and building a detailed representation of a natural environment in an online fashion. While simulated results demonstrate an ATE of 0.6m after loop closing, experimental ones confirm the system's stability in the condition of cluttered surroundings in riverine environment. This opens up various applications of SLAM in many robotic tasks: online path planning, obstacle avoidance, etc.

## ACKNOWLEDGMENT

We acknowledge the support of time and facilities from Laboratory of Advance Design and Manufacturing Processes, Ho Chi Minh City University of Technology (HCMUT), VNU-HCM for this study.

## REFERENCES

[1] C. Cadena, L. Carlone, H. Carrillo, Y. Latif, D. Scaramuzza, J. Neira and I. Reid, "Past, Present, and Future of Simultaneous Localization and Mapping: Toward the Robust-Perception Age," *IEEE Transactions on Robotics,* vol. 32, no. 6, pp. 1309-1332, 2016.

[2] W. Hess, D. Kohler, H. Rapp and D. Andor, "Real-time loop closure in 2D LIDAR SLAM," in *2016 IEEE International Conference on Robotics and Automation (ICRA)*, Stockholm, 2016.

[3] X. Chen, A. Milioto, E. Palazzolo, P. Giguère, J. Behley and C. Stachniss, "SuMa++: Efficient LiDAR-based Semantic SLAM," in *2019 IEEE/RSJ International Conference on Intelligent Robots and Systems (IROS)*, Macau, 2019.

[4] J. Han, Y. Cho and J. Kim, "Coastal SLAM With Marine Radar for USV Operation in GPS-Restricted Situations," *IEEE Journal of Oceanic Engineering,* vol. 44, no. 2, pp. 300-309, 2019.

[5] S. Rahman, A. Q. Li and I. Rekleitis, "SVIn2: An Underwater SLAM System using Sonar, Visual, Inertial, and Depth Sensor," in *2019 IEEE/RSJ International Conference on Intelligent Robots and Systems (IROS)*, Macau, 2019.

[6] R. Mur-Artal, J. M. M. Montiel and J. D. Tardós, "ORB-SLAM: A Versatile and Accurate Monocular SLAM System," *IEEE Transactions on Robotics,* vol. 31, no. 5, pp. 1147-1163, 2015.

[7] G. Grisetti, C. Stachniss and W. Burgard, "Improved Techniques for Grid Mapping With Rao-Blackwellized Particle Filters," *IEEE Transactions on Robotics,* vol. 23, no. 1, pp. 34-46, 2007.

[8] S. Kohlbrecher, O. v. Stryk, J. Meyer and U. Klingauf, "A flexible and scalable SLAM system with full 3D motion estimation," in *2011 IEEE International Symposium on Safety, Security, and Rescue Robotics*, Kyoto, 2011.

[9] M. Miller, S.-J. Chung and S. Hutchinson, "The Visual–Inertial Canoe Dataset," *The International Journal of Robotics Research,* vol. 37, no. 1, pp. 13-20, 2018.

[10] S. Griffith, G. Chahine and C. Pradalier, "Symphony Lake Dataset," *International Journal of Robotics Research,* vol. 36, no. 11, p. 1151–1158, 2017.

[11] X. Wu and C. Pradalier, "Illumination Robust Monocular Direct Visual Odometry for Outdoor Environment Mapping," in *2019 International Conference on Robotics and Automation (ICRA)*, Montreal, 2019.

[12] P. Agarwal, G. D. Tipaldi, L. Spinello and W. Burgard, "Robust map optimization using dynamic covariance scaling," in *2013 IEEE International Conference on Robotics and Automation*, Karlsruhe, 2013.

[13] J. Engel, J. Stückler and D. Cremers, "Large-scale direct SLAM with stereo cameras," in *2015 IEEE/RSJ International Conference on Intelligent Robots and Systems (IROS)*, Hamburg, 2015.

[14] E. B. Olson, "Real-time correlative scan matching," in *2009 IEEE International Conference on Robotics and Automation*, Kobe, 2009.

[15] E. Olson, "M3RSM: Many-to-many multi-resolution scan matching," in *2015 IEEE International Conference on Robotics and Automation (ICRA)*, Seattle, 2015.

[16] A. Censi, "An ICP variant using a point-to-line metric," in *2008 IEEE International Conference on Robotics and Automation*, Pasadena, CA, 2008.

[17] A. Censi, "An accurate closed-form estimate of ICP's covariance," in *Proceedings 2007 IEEE International Conference on Robotics and Automation*, Roma, 2007.

[18] Z. Zhang, "Parameter estimation techniques: A tutorial with application to conic fitting," *Image and Vision Computing,* vol. 15, no. 1, pp. 59-76, 1997.

[19] J. T. Barron, "A General and Adaptive Robust Loss Function," in *2019 IEEE/CVF Conference on Computer Vision and Pattern Recognition (CVPR)*, Long Beach, CA, USA, 2019.

[20] S. Thrun, W. Burgard and D. Fox, Probabilistic Robotics, The MIT Press, 2006, pp. 284-286.

[21] J. E. Bresenham, "Algorithm for computer control of a digital plotter," *IBM Systems Journal,* vol. 4, no. 1, pp. 25-30, 1965.

[22] R. Kümmerle, G. Grisetti, H. Strasdat, K. Konolige and W. Burgard, "G2o: A general framework for graph optimization," in *2011 IEEE International Conference on Robotics and Automation*, Shanghai, 2011.

[23] J. Sturm, N. Engelhard, F. Endres, W. Burgard and D. Cremers, "A benchmark for the evaluation of RGB-D SLAM systems," in *2012 IEEE/RSJ International Conference on Intelligent Robots and Systems*, Vilamoura, 2012.